%% file: main.tex
\documentclass[10pt,twocolumn,letterpaper]{article}

\usepackage[pagenumbers]{iccv} 


\definecolor{iccvblue}{rgb}{0.21,0.49,0.74}
\usepackage[pagebackref,breaklinks,colorlinks,allcolors=iccvblue]{hyperref}

\usepackage{graphicx}
\usepackage{array}
\usepackage[table]{xcolor}
\usepackage{booktabs}
\usepackage{bm}

\definecolor{lightyellow}{RGB}{182, 215, 242}
\definecolor{lightgray}{RGB}{251,  227, 213}
\definecolor{mplanecolor}{HTML}{FF7F7F}
\definecolor{hexplanecolor}{HTML}{ADD8E6}

\def\paperID{} 
\def\confName{CVPR}
\def\confYear{2026}

\title{\texttt{EvoGS}: 4D Gaussian Splatting as a Learned Dynamical System}

\author{
  Arnold Caleb Asiimwe \\
  Princeton University \\
  {\tt\small asiimwe@cs.princeton.edu}
  \and
  Carl Vondrick \\
  Columbia University \\
  {\tt\small vondrick@cs.columbia.edu}
}

\begin{document}
\maketitle

\vspace{-10pt}

\input{sec/0_abstract}    
\input{sec/1_intro}
\input{sec/2_relatedwork}
\input{sec/3_method}

\input{sec/4_results}
\input{sec/5_discussion}

\input{sec/6_conclusion}

{
    \small
    \bibliographystyle{ieeenat_fullname}
    \bibliography{main}
}

\end{document}

%% file: sec/0_abstract.tex
\begin{abstract}
We reinterpret 4D Gaussian Splatting as a continuous-time dynamical system, where scene motion arises from integrating a learned neural dynamical field rather than applying per-frame deformations. This formulation, which we call \textbf{\texttt{EvoGS}}, treats the Gaussian representation as an evolving physical system whose state evolves continuously under a learned motion law. This unlocks capabilities absent in deformation-based approaches: (1) sample-efficient learning from sparse temporal supervision by modeling the underlying motion law; (2) temporal extrapolation enabling forward and backward prediction beyond observed time ranges; and (3) compositional dynamics that allow localized dynamics injection for controllable scene synthesis. Experiments on dynamic scene benchmarks show that \textbf{\texttt{EvoGS}} achieves better motion coherence and temporal consistency compared to deformation-field baselines while maintaining real-time rendering.\footnote{Project page: \url{https://arnold-caleb.github.io/evogs}.}
\end{abstract}

%% file: sec/1_intro.tex
\section{Introduction}
\label{sec:intro}

\begin{figure}[t]
\centering
\includegraphics[width=\linewidth]{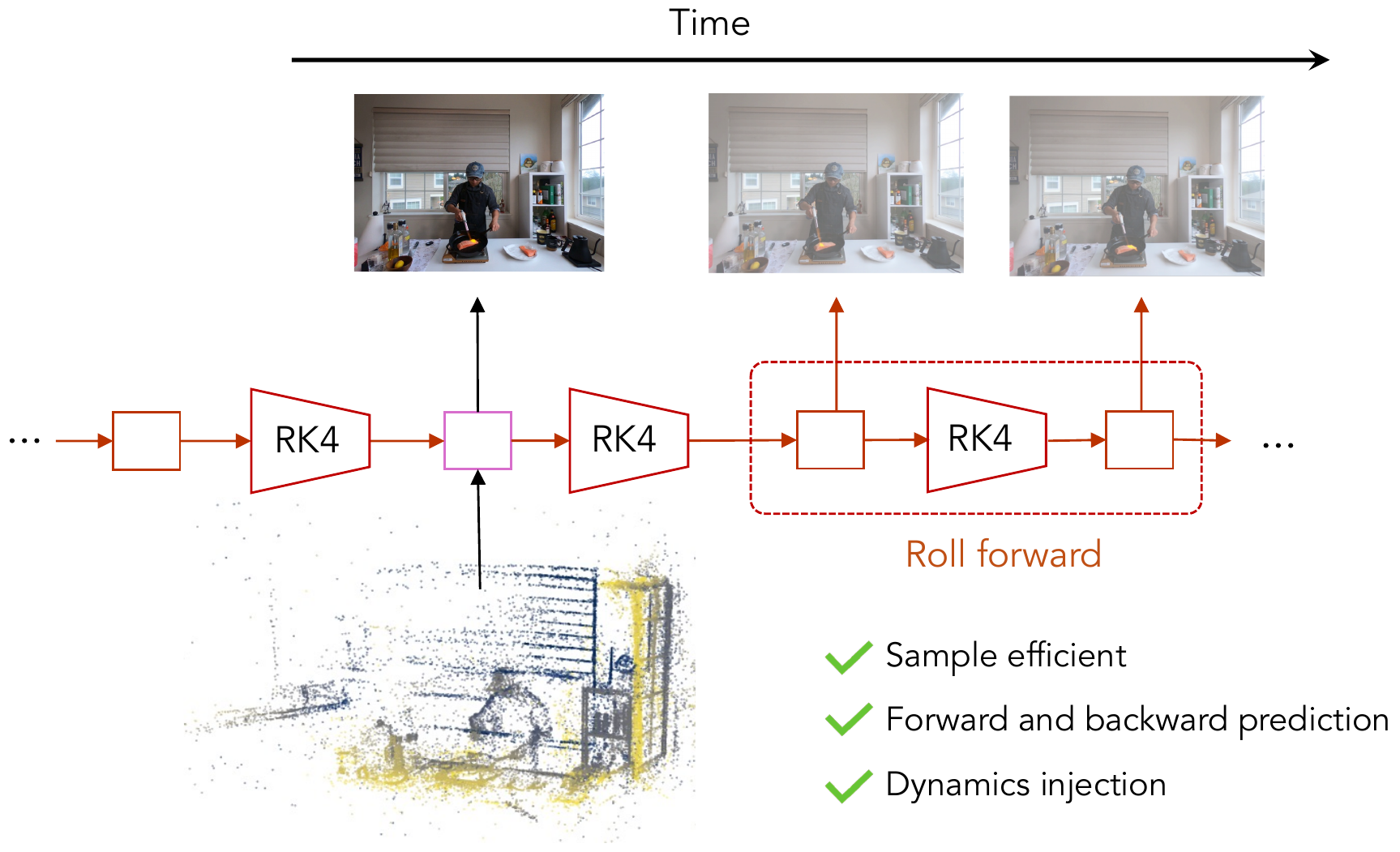}
\caption{\texttt{EvoGS} learns a continuous-time dynamical system that governs the evolution of Gaussian primitives. A neural velocity field $v_\theta$ drives their motion through numerical integration. Unlike discrete deformation-based approaches (Fig.~\ref{fig: motivation}), \texttt{EvoGS} reconstructs unseen timesteps by following the learned dynamics, enabling continuous-time extrapolation and controllable motion composition.}
\label{fig:teaser}
\vspace{-5mm}
\end{figure}

\cref{fig:teaser} \textit{``Everything flows"}---Heraclitus~\cite{kirk1983presocratic}  \\

Dynamic scene reconstruction has traditionally focused on recovering time-varying geometry and appearance from video. While early progress was driven by dynamic extensions of NeRF~\cite{mildenhall2020nerf}, these approaches rely on learned deformation fields that warp a canonical scene to each timestep~\cite{park2021nerfies, pumarola2021d, li2022neural, park2021hypernerf}. Although conceptually elegant, deformation-based NeRFs require dense and regular frame sampling, and their deformation fields often collapse when supervision becomes sparse or irregular. They are also computationally costly, as every frame requires evaluating both the canonical radiance field and its deformation.

To improve scalability and stability, subsequent works represent time as an explicit axis in a factorized 4D grid~\cite{kplanes_2023, Cao2023HEXPLANE, Chen22_Tensorf}, enabling faster, more robust rendering. However, these grid-based models still treat time as a discrete index and therefore cannot reason over missing frames or extrapolate beyond the observed temporal window. Their motion representation is descriptive rather than predictive.

Building upon explicit representations, recent advances in 4D Gaussian Splatting
\cite{Kerbl2023_3DGaussianSplatting, Wu2024_4DGaussianSplatting, lin2024gaussian, feng2024gaussian, yang2023deformable3dgs}
model dynamic scenes by updating Gaussian parameters at discrete timestamps.
While these approaches differ in how the updates are predicted—ranging from
independent per-frame Gaussian clouds \cite{luiten2023dynamic}
to framewise deformation fields \cite{yang2023deformable3dgs, duisterhof2024deformgs, huang2023sc, Wu2024_4DGaussianSplatting} via a canonical-to-world mapping (\cref{fig: motivation})—
they all share the same discrete-time assumption: motion is represented only at the observed frames.
As a result, they struggle to maintain coherent trajectories when temporal observations are sparse, irregular,
or missing entirely.

Unfortunately, partial observability is the norm outside controlled lab environments. Real-world video streams suffer from missing frames due to camera outages, irregular motion-capture sessions, rolling-shutter artifacts, and dropped frames caused by unreliable networks. In such settings, discrete deformation-based methods often fail to maintain physically meaningful trajectories at unseen timesteps: NeRF variants may freeze or produce ghost artifacts , while 4D Gaussian methods can smoothly interpolate yet drift away from the correct motion path (\cref{fig: dnerf_result}).

This reveals a fundamental limitation: \textbf{\textit{when time is discretized, models cannot robustly interpolate missing frames or reliably predict future ones.}} Yet both capabilities are crucial. Robust interpolation enables faithful reconstruction under sparse temporal observations, and the ability to predict future motion opens the door to high-stakes applications where anticipating outcomes—such as potential collisions or system failures—can prevent catastrophic events. To address these shortcomings, we propose to reinterpret dynamic scene modeling through the lens of \textbf{continuous-time dynamical systems} rather than discrete collections of warped frames. In our formulation, each Gaussian primitive behaves like a particle governed by an underlying velocity field $v_\theta(\mathbf{x}, t)$. Rather than predicting per-frame displacements, the model learns this velocity field directly, and Gaussian parameters evolve through numerical integration (Fig.~\ref{fig:teaser}). This allows the scene to be rendered at any continuous moment—including frames that were unobserved during training or timesteps far beyond the original video. We call this framework \texttt{\textbf{EvoGS}}.

By treating 4D Gaussian splatting as a learned dynamical system, \texttt{EvoGS} inherits the rendering efficiency of explicit Gaussians while enabling capabilities absent in prior work: Sparse temporal reconstruction (\S\ref{subsec:setup}): \texttt{EvoGS} learns coherent motion from as little as one-third of the total frames. Future and past prediction: Continuous integration supports  extrapolation for simulating unseen motion. Compositional motion editing: The learned velocity field enables blending, injecting, or modulating local dynamics (§\ref{subsec:compositional}).

Conceptually, EvoGS echoes ideas from dynamical systems, neural ODEs~\cite{Chen2018NeuralODE}, and filtering-based models~\cite{kalman1960new, kalman1961new} and combines prediction from continuous dynamics, correction from observations, and stabilization from temporal consistency priors. This yields coherent scene evolution even under sparse supervision and enables reliable reconstruction and prediction beyond the capabilities of existing deformation-based methods.

%% file: sec/2_relatedwork.tex
\section{Related Work}
\label{sec:related_work}

We review three areas that inform our approach. 
Sec.~\ref{sec:dynamical_formulations} covers continuous-time dynamical formulations that motivate viewing scene evolution through learned velocity fields. 
Sec.~\ref{sec:dynamic_scene_representations} surveys dynamic neural scene representations, and Sec.~\ref{sec:dynamical_gaussian_methods} discusses recent Gaussian approaches incorporating motion priors or learned dynamics.

\subsection{Dynamical Formulations}
\label{sec:dynamical_formulations}

Modeling time-varying physical systems has a long history in computer graphics and physics-based simulation, from early elastically deformable models~\cite{Terzopoulos1987_ElasticallyDeformableModels} to classical fluid solvers~\cite{Stam1999StableFluids, Bridson2008_FluidSimulation}. More recent work incorporates differentiable physics and learning-based surrogates, enabling neural networks to approximate or constrain physical dynamics~\cite{Gregson2014_CaptureFluid, Okabe2015_Fluid, Eckert2019_ScalarFlow, Franz2021_GlobalNeural, Kim2019_DeepFluids, Wiewel2019_LatentSpaceFluids}.

Recent approaches~\cite{FluidNexus2024, Deng2023_NeuralFlowMaps} combine differentiable rendering with physics-driven simulation to reconstruct or predict fluid motion directly from video, reflecting a shift toward neural dynamical systems that jointly model perception, geometry, and motion. These ideas align with methods that approximate continuous evolution through learned velocity fields rather than discrete timesteps—most notably neural ODEs~\cite{Chen2018NeuralODE}. Within the broader context of physics-informed learning, further works demonstrate how learned surrogates can accelerate fluid simulation~\cite{Kochkov2021_MLAcceleratedCFD}, how differentiable solvers enable gradient-based reconstruction of fluid phenomena from imagery~\cite{Schenck2018_SPNets_DifferentiableFluidDynamics}, and how PDE-constrained neural networks infer motion from sparse observations~\cite{ElHassan2025_PINNsFluidDynamics}.

\begin{figure}
   \centering
   \includegraphics[width=0.9\linewidth]{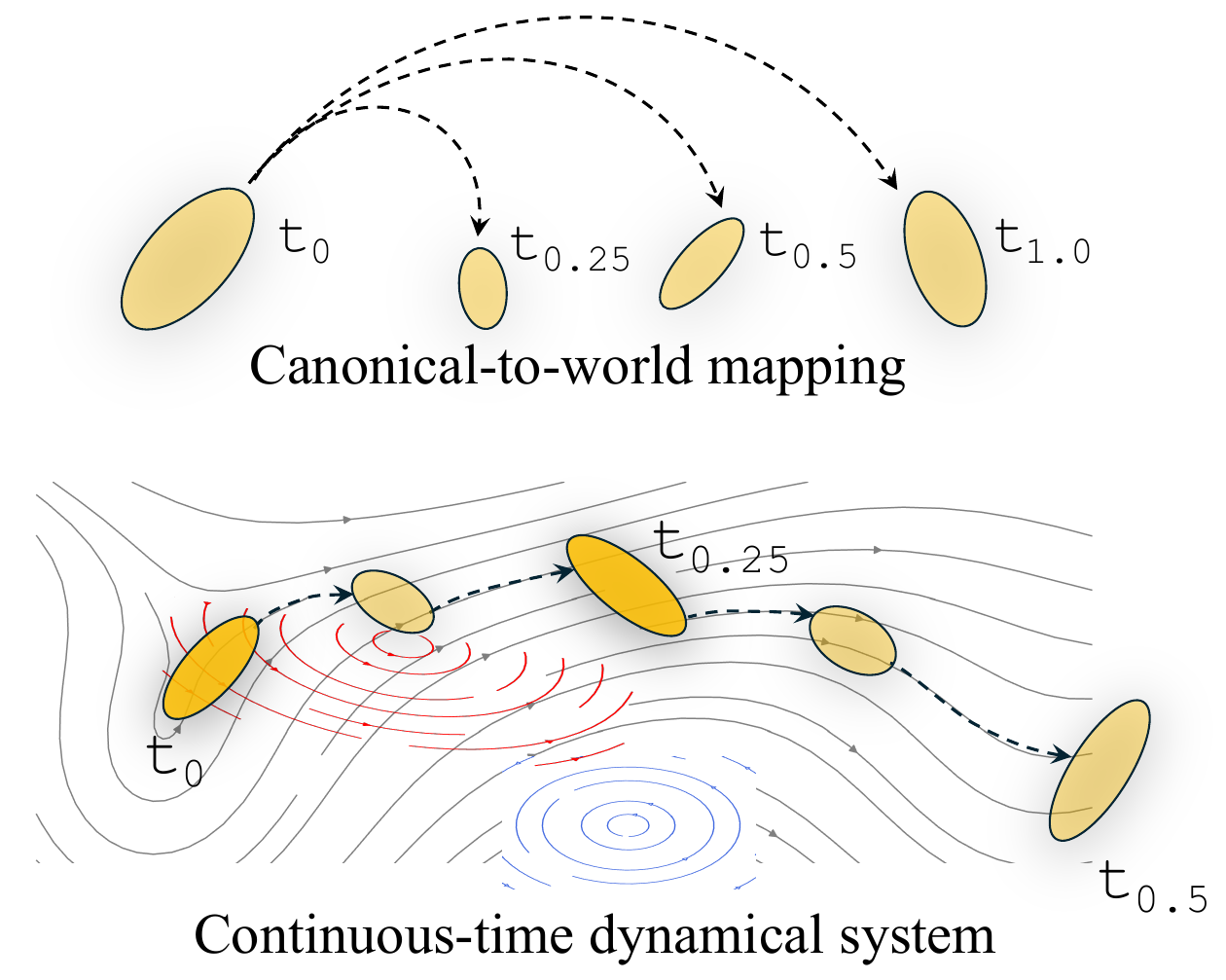}
   \caption{\textbf{Top:} Canonical deformation methods assign each timestamp an independent mapping from a shared canonical space to produce a set of per-frame transformations (learn what the scene looks like at each time $t$). \textbf{Bottom:} \texttt{EvoGS} instead learns a continuous velocity field that governs Gaussian evolution through time. Dynamics arise from integrating this field to produce reversible trajectories and coherent motion between arbitrarily spaced timestamps. The swirling field visualization shows how local dynamical structure emerges and how injected motion (\textcolor{hexplanecolor}{blue} and \textcolor{mplanecolor}{red}) blends into the learned global flow.}
   \label{fig: motivation}
   \vspace{-10pt}
\end{figure}

 \begin{figure*}[t] 
    \centering
    \includegraphics[width=\linewidth]{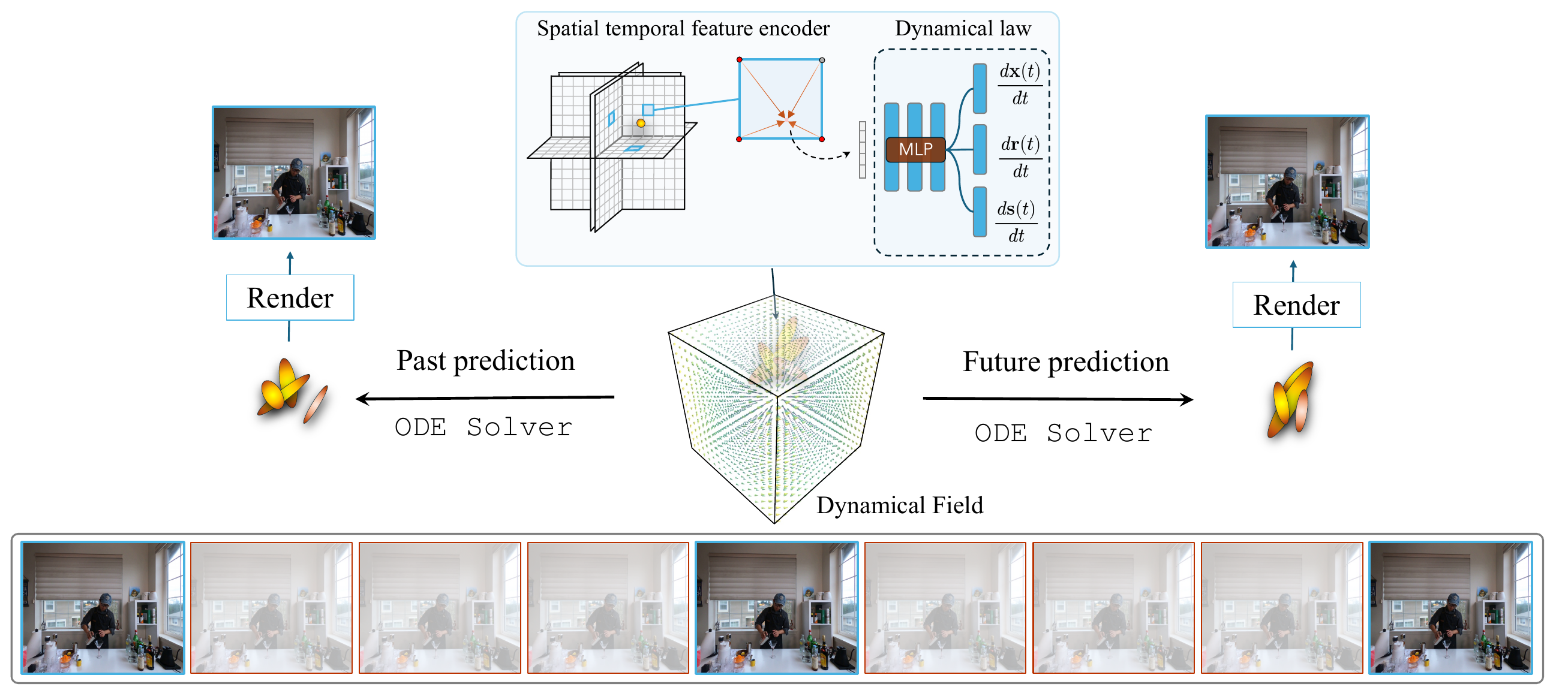}
    \caption{\textbf{Overview of \texttt{EvoGS}:} Given input frames (\textcolor{hexplanecolor}{blue}) with photometric supervision, each Gaussian is embedded using 4D spatiotemporal features and evolved through a learned continuous-time velocity field. A neural dynamical law predicts time derivatives of Gaussian attributes, and an ODE solver integrates these dynamics forward or backward to produce unseen future and past states (\textcolor{mplanecolor}{red}), which arise purely from continuous-time evolution.
    }
    \label{fig:method}
    \vspace{-3mm}
\end{figure*}

\subsection{Dynamic Scene Representations}
\label{sec:dynamic_scene_representations}

The introduction of 3D Gaussian Splatting~\cite{kerbl20233d} marked a shift from implicit neural fields~\cite{nerf, Park21_HyperNeRF, park2021nerfies} to explicit, differentiable point-based primitives for radiance field rendering. By representing a scene as a collection of anisotropic Gaussians with learnable position, orientation, opacity, and color, these methods achieve high-fidelity results through differentiable rasterization rather than volumetric integration. Their efficiency and photorealistic quality have established Gaussian splatting as a leading paradigm for explicit neural scene representation.

Extending Gaussian splatting to dynamic scenes requires modeling how Gaussian parameters evolve through time while maintaining temporal coherence and rendering efficiency. Most formulations have generalized static Gaussians into the spatiotemporal domain by learning per-frame transformations of a canonical configuration, effectively treating each timestep as an independent deformation of the scene~\cite{yang2023deformable3dgs, duisterhof2024deformgs, huang2023sc, Wu2024_4DGaussianSplatting}.
Subsequent approaches introduced temporally shared Gaussian attributes to improve coherence~\cite{Zhang2024_MEGA}, surfel-based deformation models for finer control of local motion and geometry~\cite{luiten2024dynamic}, and disentangled or editable formulations that separate static and dynamic components or apply segmentation-based priors for controllable motion~\cite{kwon2025efficient, lee2024fully}.

\subsection{Dynamical Gaussian  Methods}
\label{sec:dynamical_gaussian_methods}

Recent extensions of Gaussian splatting have introduced explicit motion modeling and learned dynamics, moving beyond frame-wise deformations---several works extend static 3D Gaussians to dynamic settings through temporally coupled transformations, motion-aware attributes, or latent motion factorization~\cite{lee2024fully, Kratimenos2024_DynMF, Guo2024_MotionAware3DGS, Hu2025_MotionDecoupled3DGS}. Others focus on motion guidance and continuous motion cues to handle large or blurred motions~\cite{Zhu2024_MotionGS, Lee2025_CoMoGaussian}. Inspired by physical systems, some approaches embed motion laws within Gaussian primitives, treating each as a particle evolved under continuum or flow-based dynamics~\cite{Xie_2024_CVPR}. Other formulations express temporal variations—such as position or covariance—as compact parametric functions of time, e.g., polynomial or Fourier expansions~\cite{Lin2024GaussianFlow}. Self-supervised variants further learn scene flow for dynamic or unlabeled environments~\cite{SplatFlow2024}.

Despite these advances, existing methods still rely on discrete temporal updates or per-frame optimization, requiring dense supervision and struggling to extrapolate motion beyond observed frames. In contrast, our approach models Gaussian evolution as a continuous-time process governed by a neural velocity field $v_\theta(\mathbf{x},t)$, enabling controllable motion composition, sparse-frame training, and temporally coherent rollouts.

%% file: sec/3_method.tex
\section{Method}
\label{sec:method}

This section introduces \texttt{EvoGS} (\cref{fig:method}), a continuous-time formulation of dynamic Gaussian splatting.
We first outline the model design (Sec.\ref{sec:method_overview}), then describe the feature encoder (Sec.\ref{sec:method_encoder}),
the neural dynamical law (Sec.\ref{sec:method_dynamics}), and a Kalman-inspired stabilization mechanism (Sec.\ref{sec:method_waypoints}).
We conclude with the rendering process and training objective (Sec.\ref{sec:method_render}).

\subsection{Overview}
\label{sec:method_overview}

We treat each Gaussian as a particle evolving under a learned continuous-time dynamical system.
Its trajectory is defined by a neural velocity field conditioned on local spatiotemporal features. Following standard practice in Gaussian Splatting~\cite{kerbl20233d}, all Gaussians are initialized from a point cloud reconstructed via structure-from-motion (SfM). \texttt{EvoGS} (Fig.~\ref{fig:method}) then consists of:
(1) a 4D feature encoder that produces local embeddings from a factorized space–time representation (e.g., HexPlane~\cite{Cao2023HEXPLANE}),
(2) a neural dynamical law predicting instantaneous time derivatives of Gaussian attributes, and
(3) a differentiable ODE integrator that advances these states in time.

\subsection{Spatiotemporal Feature Encoding}
\label{sec:method_encoder}

Each Gaussian center $\mathbf{p}_i=(x_i,y_i,z_i)$ at time $t$ is embedded via bilinear interpolation over six 
space–time factorization planes
$\{\mathbf{P}_{xy},\mathbf{P}_{xz},\mathbf{P}_{yz},\mathbf{P}_{xt},\mathbf{P}_{yt},\mathbf{P}_{zt}\}$:
\[
\mathbf{f}_i(t)=\Phi(\mathbf{p}_i,t),
\]
where $\Phi$ denotes the differentiable lookup from the 4D grid.
These features encode local geometry and motion cues and condition the velocity field used in the dynamical update.

\subsection{Neural Dynamical Law}
\label{sec:method_dynamics}

Each Gaussian primitive has a state
\[
\mathbf{x}_i(t)
=
[\mathbf{p}_i(t),\, \mathbf{R}_i(t),\, \mathbf{S}_i(t),\, \mathbf{c}_i(t),\, \alpha_i(t)],
\]
where $\mathbf{p}_i$ is its 3D position, $\mathbf{R}_i$ its rotation (parameterized via an exponential-map
update), $\mathbf{S}_i$ its anisotropic scale, $\mathbf{c}_i$ its color, and $\alpha_i$ its opacity. The state evolves according to the continuous-time ODE
\begin{equation}
\frac{d\mathbf{x}_i}{dt}
= \mathbf{v}_\theta(\mathbf{x}_i(t),\,\mathbf{f}_i(t),\,t),
\label{eq:dynamics}
\end{equation}
where $\mathbf{v}_\theta$ is a lightweight MLP predicting derivatives of position, rotation, and scale.
We integrate this ODE with a differentiable solver (RK4), enabling both forward and backward
temporal propagation:
\begin{equation}
\begin{aligned}
\mathbf{x}_i(t_1)
&= \mathrm{RK4}(\mathbf{x}_i(t_0),\, t_0,\, \Delta t,\, \mathbf{v}_\theta),\\
\mathbf{x}_i(t_0)
&= \mathrm{RK4}(\mathbf{x}_i(t_1),\, t_1,\,-\Delta t,\, \mathbf{v}_\theta).
\end{aligned}
\label{eq:bidirectional}
\end{equation}
Bidirectional integration yields reversible dynamics and allows the model to propagate motion through
missing frames or ambiguously observed regions.

\begin{figure*}
   \centering
   \includegraphics[width=1.0\linewidth]{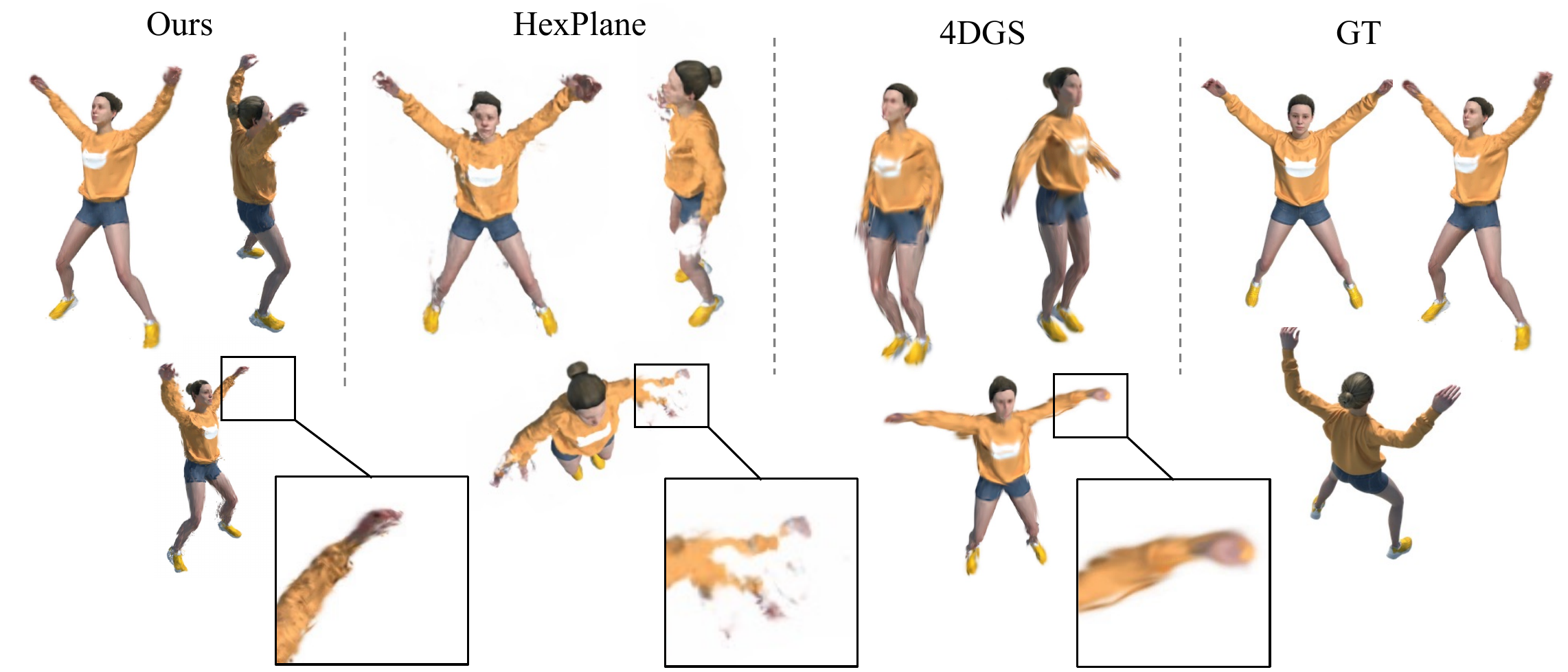}
   \caption{Comparison of \texttt{EvoGS} on reconstruction of unseen dynamic human motion on the Jumping Jacks scene.
Compared to HexPlane~\cite{Cao2023HEXPLANE} and 4DGS~\cite{Wu2024_4DGaussianSplatting}, which breakdown for unseen timesteps (e.g., limbs rupturing or blurring)}
      \label{fig: dnerf_result_go}
      \vspace{-10pt}
\end{figure*}

\subsection{Gaussian Waypoints for Motion Stabilization}
\label{sec:method_waypoints}

Continuous ODE integration can accumulate drift over long temporal horizons due to numerical error and locally underconstrained motion. 
In classical filtering, such drift is controlled by alternating prediction and correction steps. 
While a full Kalman filter is infeasible here---given nonlinear dynamics, millions of latent states, and non-Gaussian rendering losses---we adopt a related idea using \emph{Gaussian waypoints}. During training, a small number of anchor snapshots $\mathcal{A}=\{t^{(a)}_1, t^{(a)}_2, \ldots\}$  store the Gaussian states at fixed times. These anchors act as sparse pseudo-observations of the underlying dynamical system.

For any target frame at time $t$, we locate the nearest past anchor $t^{(a)}$ and 
reinitialize the ODE state using the stored Gaussian parameters at $t^{(a)}$, 
then integrate forward from $t^{(a)}$ to $t$. That way, the effective integration horizon is reduces so that drift accumulation is reduced and prevents diverging during long rollouts.

Optionally, we penalize deviations between the integrated state and the stored anchor snapshot itself:
\[
\mathcal{L}_{\text{anchor}}
=
\sum_{t^{(a)} \in \mathcal{A}}
\|\mathbf{x}(t^{(a)}) - \hat{\mathbf{x}}(t^{(a)})\|_2^2,
\]
where $\hat{\mathbf{x}}(t^{(a)})$ is the anchor state and $\mathbf{x}(t^{(a)})$ is the state obtained by integrating from the preceding anchor.
This encourages consistency with anchor waypoints while still allowing smooth continuous-time evolution between them. In contrast to classical filters, we do not maintain explicit velocity estimates or covariance; the anchors function solely as sparse, fixed reference states that constrain long-term integration.

\subsection{Rendering and Objective}
\label{sec:method_render}

At each target timestamp $t_1$, the evolved Gaussians $\mathcal{G}(t_1)$ are rendered using differentiable
Gaussian splatting~\cite{kerbl20233d}. Supervision is provided by a standard photometric reconstruction loss
(L1, optionally combined with SSIM/LPIPS).

To encourage stable motion and suppress drift, we include temporal smoothness on the spatiotemporal planes
(plane TV and time-smoothing), as well as a velocity-coherence regularizer to encourage nearby Gaussians to move consistently. When anchor waypoints are enabled, we apply a soft anchor-consistency term
that pulls integrated states toward stored anchor snapshots.

The full training objective is:
\begin{equation}
\mathcal{L}
=
\mathcal{L}_{\text{photo}}
+ \lambda_{\text{coh}} \mathcal{L}_{\text{coh}}
+ \lambda_{\text{anchor}} \mathcal{L}_{\text{anchor}}
+ \lambda_{\text{tv}} \mathcal{L}_{\text{tv}},
\label{eq:final_loss}
\end{equation}
where $\mathcal{L}_{\text{coh}}$ enforces velocity coherence,  
$\mathcal{L}_{\text{anchor}}$ applies the optional anchor constraint,  
and $\mathcal{L}_{\text{tv}}$ smooths the spatiotemporal feature fields.

\begin{figure*}
   \centering
   \includegraphics[width=1.0\linewidth]{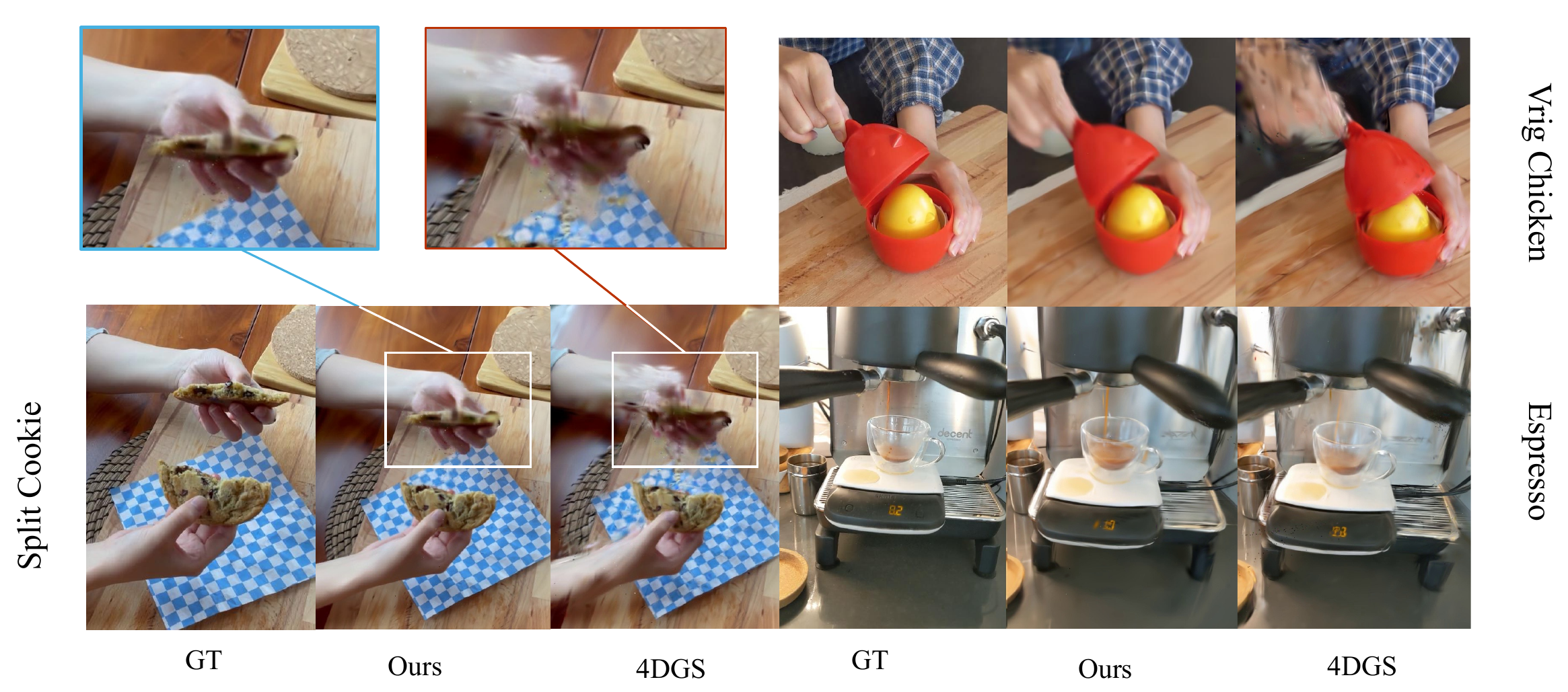}
   \caption{\textbf{Extrapolation on real monocular dynamic scenes.} Comparison on Split Cookie, Vrig Chicken, and Espresso sequences, where the model must predict frames beyond the observed time range. We include comparisons to ~\cite{yang2023deformable3dgs, Pumarola21_DNeRF, kplanes_2023} in suppl. for completeness.}
      \label{fig: dnerf_result_first}
      \vspace{-10pt}
\end{figure*}

\begin{figure}
   \centering
   \includegraphics[width=1.0\linewidth]{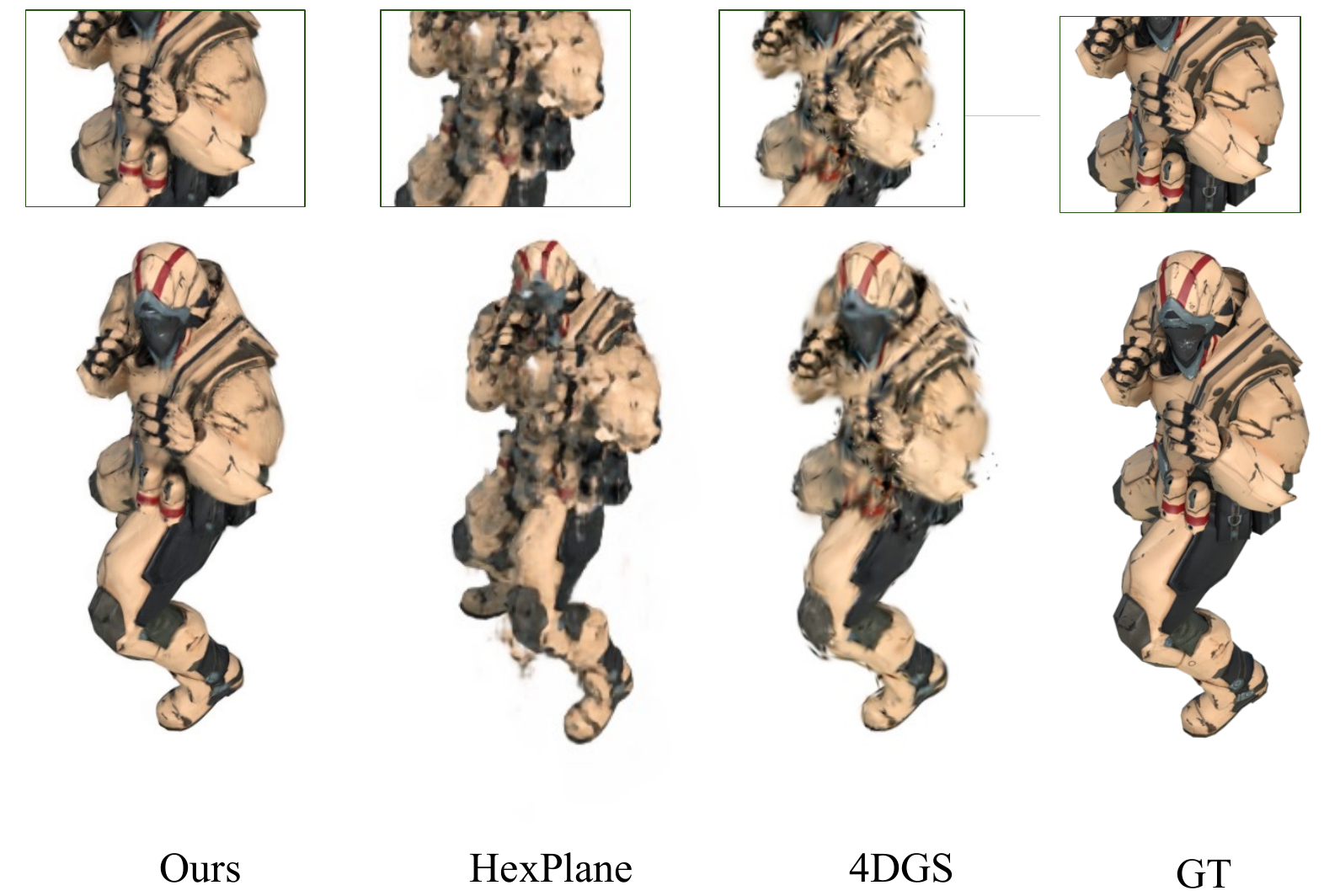}
   \caption{\textbf{Interpolation on the Hook scene.}
    \texttt{EvoGS} maintains coherent geometry and motion, whereas HexPlane freezes the dynamics and 4DGS produces an over-smoothed intermediate frame.}
      \label{fig: dnerf_result}
      \vspace{-10pt}
\end{figure}

%% file: sec/4_results.tex
\section{Experiments}
\label{sec:experiments}

We evaluate \texttt{EvoGS} on synthetic and real-world datasets, comparing against state-of-the-art
dynamic scene reconstruction methods~\cite{Cao2023HEXPLANE, kplanes_2023, Wu2024_4DGaussianSplatting}. 
Section~\ref{subsec:setup} describes implementation details, datasets, and experimental settings.
Section~\ref{subsec:compositional} demonstrates external motion injection and controllable dynamics.
Section~\ref{subsec:ablations} provides ablation studies and analysis.

\subsection{Experimental Setup}
\label{subsec:setup}

\begin{figure*}[t]
    \centering
    \includegraphics[width=\linewidth]{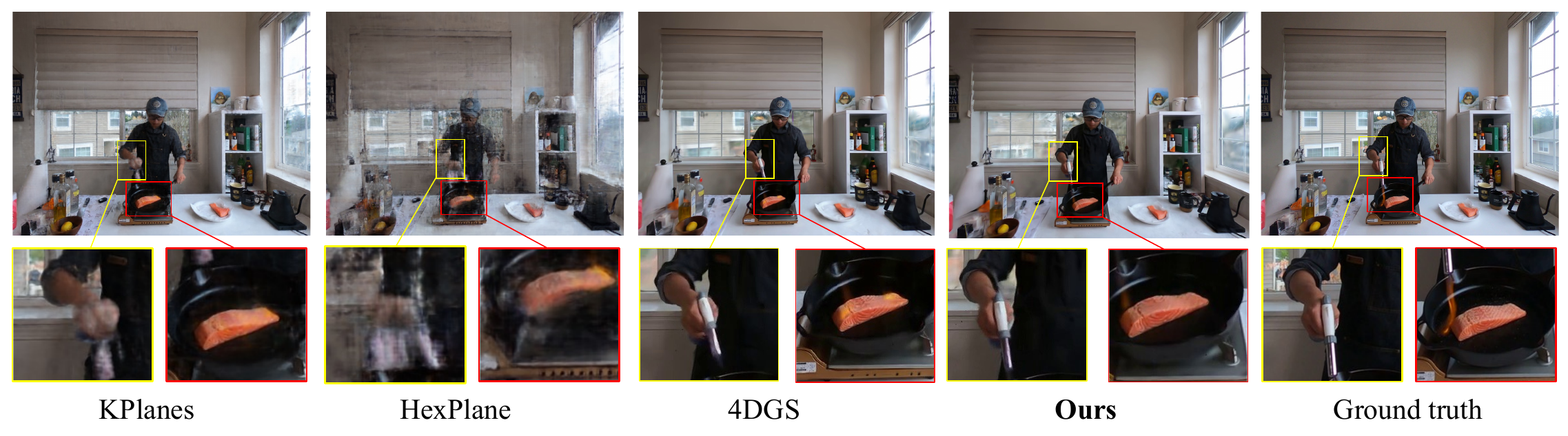}
    \caption{\textbf{Interpolation under sparse-frame training on the N3DV ``flame salmon" scene.} K-Planes~\cite{kplanes_2023} and HexPlane~\cite{Cao2023HEXPLANE} largely freeze the motion when frames are skipped, while 4DGS~\cite{Wu2024_4DGaussianSplatting} preserves appearance but smoothes fast-moving regions (e.g., the hand becomes shortened or smeared).}
    \label{fig:roll_forward}
\end{figure*}

\paragraph{Implementation Details.}
Our model is implemented in PyTorch and trained on a single NVIDIA L40 GPU.
We adopt the optimization settings of~\cite{Wu2024_4DGaussianSplatting}, with minor adjustments for
continuous-time dynamics.  
To assess temporal robustness, we primarily evaluate in the \emph{sparse-frame} regime by dropping
frames during training (Figs.~\ref{fig: dnerf_result_go},~\ref{fig: dnerf_result_first},~\ref{fig: dnerf_result},~\ref{fig:roll_forward}). The datasets used are described next.

\paragraph{Datasets.}
For synthetic evaluation, we use the D-NeRF dataset~\cite{Pumarola21_DNeRF}, which contains monocular
dynamic scenes with 50--200 frames and randomly varying camera trajectories. For real-world evaluation, we use the Neural 3D Video (N3DV) dataset~\cite{Li_2022_CVPR}, which provides
multi-view dynamic captures with calibrated poses and complex nonrigid motion, and the Nerfies
dataset~\cite{park2021nerfies}, consisting of monocular captures with moderate to fast nonrigid motion.  All experiments use the provided camera parameters. For each sequence, we uniformly subsample frames for training and evaluation, as detailed below.

\paragraph{Sparse-Frame and Extrapolation Settings.}
To evaluate temporal generalization, we train using every $k$-th frame ($k \in \{2, 4, 8, 10\}$) of
each sequence. On N3DV (300 frames), this results in only $300/k$ training frames. We report results for
strides $k=2$ and $k=8$ in Table~\ref{tab:results_table}. We also evaluate \emph{future extrapolation} (\cref{fig: dnerf_result}) by training only on the first 0.75 fraction
of frames and predicting all unseen future frames. The same protocol can be applied for \emph{backward
rollout}, where the learned velocity field is integrated backward to reconstruct earlier frames. The sparse-frame and extrapolation settings are used consistently across N3DV, D-NeRF, and Nerfies datasets. 

\paragraph{Waypoint initialization}
To reduce long-term drift during continuous integration, we introduce three temporal anchors placed at the start, midpoint, and end of each sequence. Each anchor corresponds to a 3D Gaussian state rendered at its timestamp and acts as a soft constraint that keeps the learned trajectories consistent over long time horizons.

\paragraph{Metrics.}
We evaluate reconstruction quality using standard photometric metrics: PSNR, SSIM, and LPIPS~\cite{zhang2018unreasonable}. 
All results are reported on held-out frames under the same sparse-frame or extrapolation protocols used in training.

\subsection{Compositional and Controllable Dynamics}
\label{subsec:compositional}

\begin{figure}[t] 
    \centering
    \includegraphics[width=\linewidth]{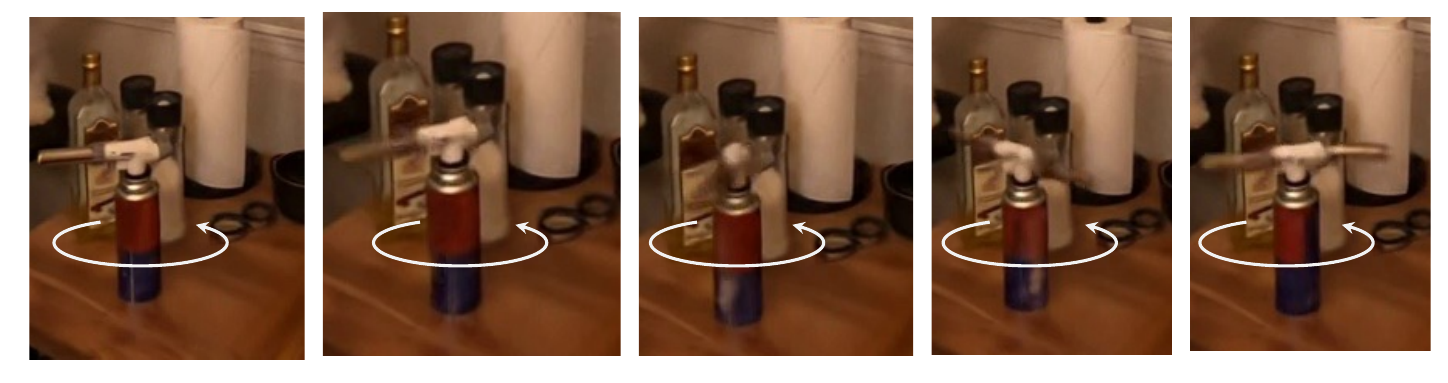}
    \caption{
    \textbf{Compositional Dynamics injection:}
    By locally blending a rotational velocity field (indicated by white arrow), \texttt{EvoGS} can inject new fields into a scene. }
    \label{fig:compositional_dynamics}
\end{figure}

A key advantage of representing scene motion as a continuous-time velocity field is enabling \emph{controllable motion synthesis}. Since dynamics are encoded as a vector field we can directly manipulate, mix, or replace portions of the flow to produce new motion without retraining (\cref{fig:compositional_dynamics}).

\paragraph{Velocity field composition and local dynamics injection.}
Formally, given two velocity fields—a learned field $\mathbf{v}_\theta$ and an external field $\mathbf{v}_{\text{ext}}$ (e.g., a user-defined motion or a field borrowed from another model)—we can form a spatially mixed field, enabling a simple \emph{vector-field algebra}:
\begin{equation}
\mathbf{v}_{\text{mix}}(\mathbf{x},t)
= \lambda(\mathbf{x})\,\mathbf{v}_\theta(\mathbf{x},t)
+ \bigl(1 - \lambda(\mathbf{x})\bigr)\,\mathbf{v}_{\text{ext}}(\mathbf{x},t).
\end{equation}

where $\lambda(\mathbf{x})\!\in\![0,1]$ is a spatial mask controlling which region follows which dynamics.
This allows selected objects to inherit new motion. Because \texttt{EvoGS} evolves Gaussians through continuous-time integration,
$\mathbf{v}_{\text{mix}}$ yields smooth spatiotemporal transitions. \cref{fig:compositional_dynamics} shows a rotational field  combined injected into the learned field via 
\begin{equation}
\mathbf{v}'(\mathbf{x},t)
=
\lambda(\mathbf{x})\,\mathbf{v}_{\text{inj}}(\mathbf{x},t)
+
(1-\lambda(\mathbf{x}))\,\mathbf{v}_\theta(\mathbf{x},t).
\end{equation} where $\mathbf{v}_{\text{inj}}$. Gaussians in the masked region follow $v_{\text{inj}}$, while the rest of the scene continues under $v_\theta$ which allows new motion to be created without retraining.



\paragraph{Object incompleteness and the need for recomposition.}
Injecting a new velocity field into a scene requires a complete 3D representation of the target object. However, the Gaussians associated with an object $\mathcal{G}_{\text{obj}}$ are often incomplete: because training cameras observe the object only from a subset of angles, large portions of its surface are undersampled or entirely missing. When the object is moved or rotated, these unseen regions become
exposed and produce severe artifacts.

\paragraph{Geometry completion and reinsertion.}
We first isolate the target object using a 3D Gaussian segmentation mask $\lambda(\mathbf{x})$~\cite{cen2023saga}. To reconstruct the missing geometry, we render segmented ground-truth images from the original camera
views and use them as input to Zero123~\cite{liu2023zero1to3} to synthesize novel viewpoints that were never observed. These real and synthesized views supervise a second-pass 3D Gaussian optimization applied only to $\mathcal{G}_{\text{obj}}$, enabling densification and completion of the object’s geometry. The refined Gaussians are then reinserted into the full scene, and the injected velocity field $\mathbf{v}_{\text{inj}}$ is applied to them during continuous-time evolution.

\input{tables/full_width}

%% file: tables/full_width.tex
\begin{table*}[t!]
    \centering
    \caption{
    \textbf{Sparse-frame training results on the N3DV ``coffee martini'' scene.}
    We evaluate reconstruction fidelity when training on every $k$-th frame (here $k{=}2,8$). We include results on training all frames for completion. $^\ddag$Full supervision results are obtained from~\cite{Wu2024_4DGaussianSplatting}.
    }
    \vspace{-1mm}
    \resizebox{0.96\textwidth}{!}{
    \begin{tabular}{p{3.4cm}p{1.5cm}p{1.5cm}p{1.5cm}p{1.5cm}p{1.5cm}p{1.5cm}p{1.5cm}p{1.5cm}p{1.5cm}}
    \toprule[2pt]
     & \multicolumn{3}{|c|} {Full supervision} & \multicolumn{3}{c|}{$k=2$} & \multicolumn{3}{c}{$k=8$} \\
    Model & PSNR$\uparrow$  & D-SSIM$\downarrow$ & LPIPS$\downarrow$ & PSNR$\uparrow$  & SSIM$\uparrow$ & LPIPS$\downarrow$ & PSNR$\uparrow$  & SSIM$\uparrow$ & LPIPS$\downarrow$ \\
    \midrule
    HexPlane$^\ddag$~\cite{Cao2023HEXPLANE} & \textbf{31.70} & \textbf{0.014} & 0.075  & 26.14 & 0.830 & 0.30 & 24.39 & 0.782 & 0.39 \\  
    KPlanes$^\ddag$~\cite{kplanes_2023} & 31.63 & - & -   & 26.28 & 0.832 & 0.29 & 24.52 & 0.785 & 0.39 \\   
    D-NeRF~\cite{Pumarola21_DNeRF} &  29.40 & 0.028 & 0.112 & 23.80 & 0.801 & 0.48 & 23.72 & 0.721 & 0.43 \\ 
    Deformable 3DGS~\cite{yang2023deformable3dgs} &  30.52 & 0.022 & 0.084 & 25.40 & 0.84 & 0.30 & 22.12 & 0.742 & 0.41 \\ 
    4DGS$^\ddag$~\cite{Wu2024_4DGaussianSplatting} &  31.15 & 0.016 & \textbf{0.049} & 27.65 & 0.878 & 0.27 & 26.45 & 0.846 & \textbf{0.22} \\  
    \midrule
    \textbf{Ours} & 30.82 & 0.022 & 0.085 & \textbf{28.25} & \textbf{0.914} & \textbf{0.20} & \textbf{26.90} & \textbf{0.870} & 0.26 \\  
    \bottomrule [2 pt]
    \end{tabular}}
    \label{tab:results_table}
\end{table*}


%% file: sec/5_discussion.tex
\subsection{Ablations and Analysis}
\label{subsec:ablations}

\begin{figure}[t] 
    \centering
    \includegraphics[width=\linewidth]{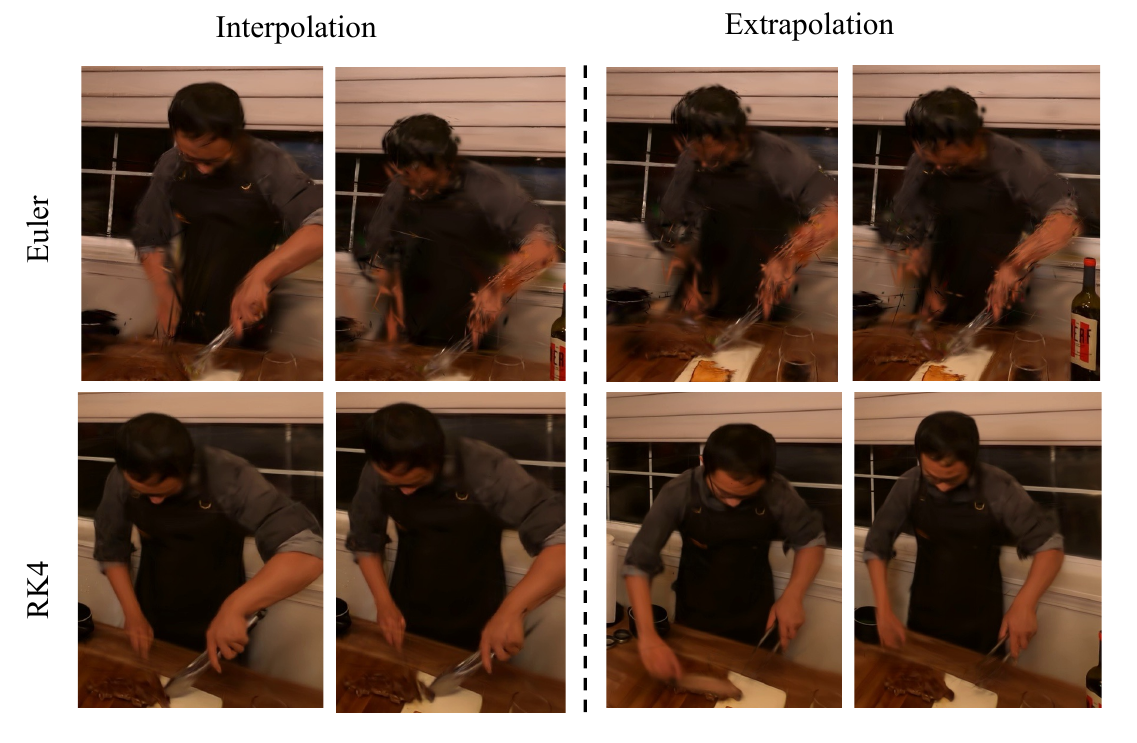}
    \caption{Euler integration rapidly accumulates temporal drift. RK4 produces smooth, consistent trajectories for both interpolation and extrapolation.}
    \label{fig:euler_vs_rk4}
    \vspace{-2mm}
\end{figure}

\begin{figure*}[t] 
    \centering
    \includegraphics[width=\linewidth]{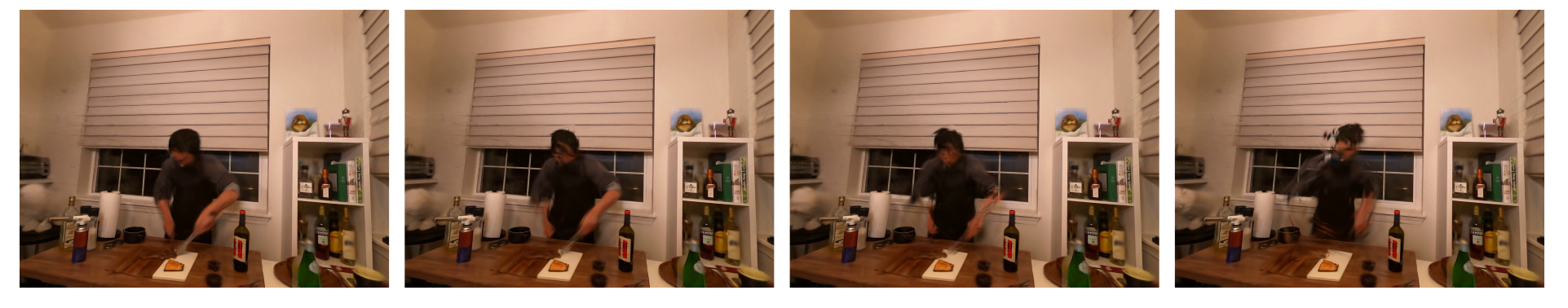}
    \caption{
        Without Gaussian waypoints, long forward integration causes rollouts to slowly drift and distort the scene
        }
    \label{fig:long_horizon}
    \vspace{-2mm}
\end{figure*}

\noindent\textbf{Integration order.}
We compare a fourth-order Runge–Kutta solver (RK4) to a first-order Euler integrator in~\cref{fig:euler_vs_rk4}.  While Euler integration is numerically cheap, it accumulates drift rapidly and incoherent motion across different gaussians as the system is rolled forward or backward in time.  RK4, by contrast, produces stable trajectories and preserves Gaussian structure, however under extremely long-horizon extrapolation the gaussian structure starts to fall apart (\cref{fig:long_horizon}).

\noindent\textbf{Effect of Gaussian waypoints.}
Removing Gaussian waypoints increases temporal drift because the ODE is integrated from a single fixed reference state and errors compound over long sequences~\cref{tab:future_reconstruction}.  Waypoints act as sparse re-initialization states: at each target timestamp the system integrates only from the nearest stored anchor, preventing the accumulation of small numerical errors.  
Without waypoints, we observe increasing trajectory divergence and noticeable spatial jitter like in \cref{fig:long_horizon}.

\noindent\textbf{Sparse-frame robustness.}
With moderate sparsity (e.g., training on every 8th frame), deformation-based and factorized spatiotemporal grids baselines struggle to infer plausible intermediate motion (\cref{fig:roll_forward}), while our continuous-time formulation maintains coherent trajectories through the learned velocity field. Under extreme sparsity (e.g., one frame every 20), the dynamics become underconstrained and the advantage over deformation-based models diminishes—both behave similarly when temporal supervision is insufficient. ~\cref{tab:results_table} summarizes performance across sparsity levels.

\section{Discussion}
We show that reconstruction and prediction can be expressed within the same continuous dynamical space. Instead of optimizing per-frame deformations, the model learns a velocity field that governs scene evolution across both observed and unobserved timestamps. This shared representation reduces temporal discontinuities and enables forward extrapolation and backward rollouts without retraining. Higher-order integration further stabilizes long-range behavior (\cref{fig:euler_vs_rk4}), suggesting that continuous-time formulations provide a strong inductive bias for modeling dynamic 3D scenes.

\input{tables/future_frames}
\begin{figure}[t] 
    \centering
    \includegraphics[width=\linewidth]{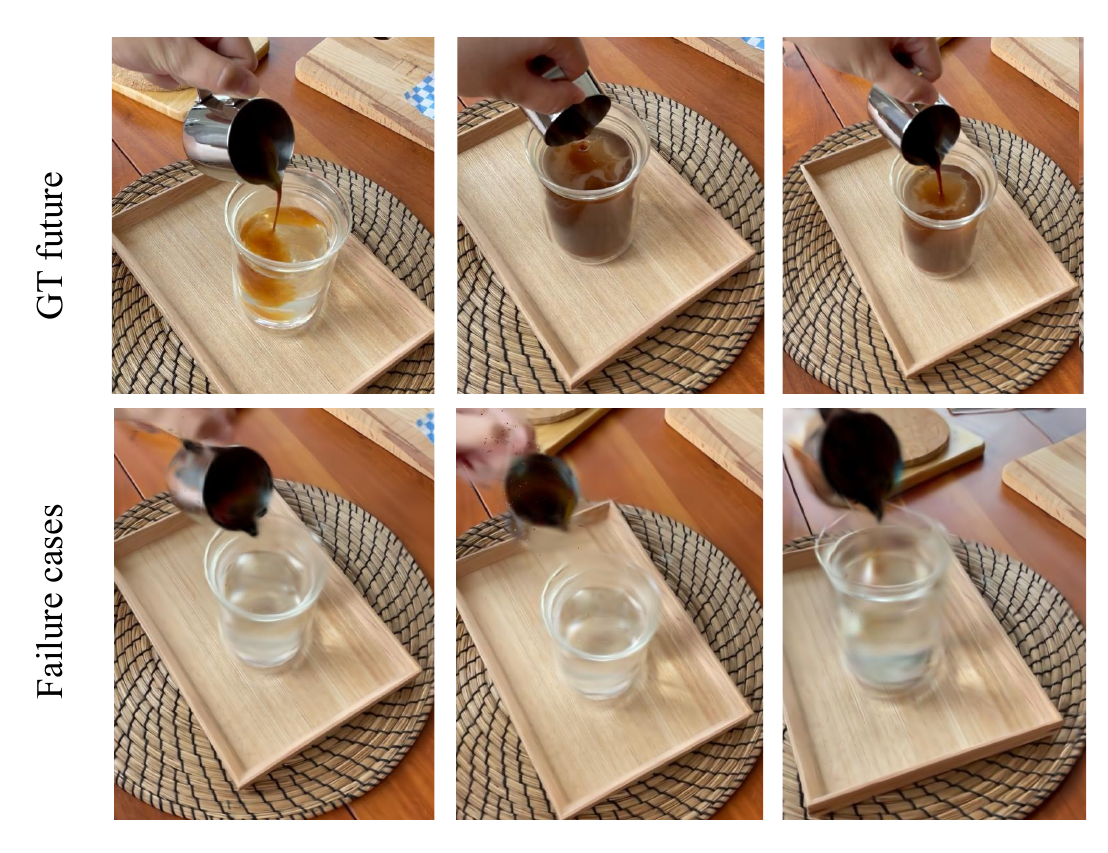}
    \caption{
        Failure case: lack of physical reasoning in emergent dynamics. When presented with scenes requiring true physical understanding—such as liquid filling a glass—\texttt{EvoGS} can extrapolate the motion of rigid objects (e.g., the hand and cup) but fails to infer the emergent fluid behavior. }
    \label{fig:failure_cases}
    \vspace{-3mm}
\end{figure}

Because motion is represented as a vector field, injecting external velocity fields provides a simple and expressive mechanism for editing 4D content.
This vector-field algebra (Sec: \ref{subsec:compositional}) enables localized motion synthesis, mixing, or replacement—all without re-optimizing the entire scene.
Such controllable dynamics hint at a broader direction: dynamic scene representations that behave like world models, in which motion rules can be modified, composed, or conditioned on external signals.

Our formulation suggests that continuous-time velocity fields may serve as a useful interface between reconstruction methods and video-generation models.
Generative world models~\cite{ha2018worldmodels, bruce2024genie, zhang2024recapturegenerativevideocamera} typically operate on latent tokens or coarse implicit grids, whereas \texttt{EvoGS} evolves explicit 3D primitives that are directly renderable.
Training such dynamical fields at larger scale—or conditioning them on text, audio, or actions—could enable generative 4D scenes with physically plausible, editable dynamics.
Dynamic Gaussian splatting may thus form a bridge between reconstruction-centric 3D methods and generative video models.

\paragraph{Limitations and opportunities.}
Our approach is data-driven and inherits the biases and ambiguities present in the training video. In scenarios requiring genuine causal or physical reasoning, the learned velocity field may fail to generalize. For example, in sequences where a hand begins to pour water into a glass (\cref{fig:failure_cases}), \texttt{EvoGS} can extrapolate the hand’s motion but cannot infer fluid behavior or anticipate water–glass interaction—phenomena that fall outside the spatiotemporal patterns observed in the training frames. Likewise, under extreme temporal sparsity, the dynamics become underconstrained and gradually regress toward deformation-like behavior.

%% file: tables/future_frames.tex
\begin{table}[t!]
    \centering
    \caption{\textbf{Ablation study on ``flame salmon scene":} Removing waypoints, coherence loss, or the HexPlane encoder degrades long-range prediction. Metrics are reported on frames held out for $t > 0.75$ (supervision on $t \le 0.25)$.}
    \vspace{-2mm}
    \resizebox{0.48\textwidth}{!}{
    \begin{tabular}{p{3.5cm} p{1.3cm} p{1.3cm} p{1.3cm}}
    \toprule[2pt]
    Model & PSNR$\uparrow$ & SSIM$\uparrow$ & LPIPS$\downarrow$ \\
    \midrule
    \textbf{Ours} (w/o $\lambda_{\text{anchor}}$)     & 24.1 & 0.846 & 0.23 \\
    \textbf{Ours} (w/o $\lambda_{\text{coh}}$)     & 25.1 & 0.872 & 0.23 \\
    \textbf{Ours} (w/o hexplane)      & 23.3 & 0.847 & 0.24 \\
    \textbf{Ours}   & \textbf{25.73} & \textbf{0.880} & \textbf{0.20} \\
    \bottomrule[2pt]
    \end{tabular}}
    \label{tab:future_reconstruction}
    \vspace{-2mm}
\end{table}

%% file: sec/6_conclusion.tex
\section{Conclusion}

We introduced \texttt{EvoGS}, a dynamic Gaussian framework that models scene evolution through a continuous-time velocity field.
Integrating Gaussian parameters over time yields a unified representation for reconstruction, interpolation, extrapolation, and controllable dynamics, without relying on per-frame deformations.
\paragraph{Acknowledgements} This research was primarily done while ACA was an undergraduate at Columbia University and completed while ACA was a graduate student at Princeton University. We thank Prof.\ Felix Heide for insightful conversations during the preparation of the paper, and the I.I.\ Rabi Scholars program at Columbia for supporting this research.


%% file: main.bib
@String(CVPR= {IEEE Conf. Comput. Vis. Pattern Recog.})

@String(ICCV= {Int. Conf. Comput. Vis.})

@String(ECCV= {Eur. Conf. Comput. Vis.})

@String(TOG= {ACM Trans. Graph.})

@String(AAAI = {AAAI})

@String(CVPR  = {CVPR})

@String(ICCV  = {ICCV})

@String(ECCV  = {ECCV})

@String(TOG   = {ACM TOG})

@book{kirk1983presocratic,
  title={The Presocratic Philosophers},
  author={Kirk, G.S. and Raven, J.E. and Schofield, M.},
  publisher={Cambridge University Press},
  year={1983},
  note={Discussion of Heraclitus’ doctrine of flux, commonly paraphrased as “everything flows”}
}

@inproceedings{Park21_HyperNeRF,
  author    = {Keunhong Park and Utkarsh Sinha and Peter Hedman and Jonathan T. Barron and Sofien Bouaziz and Dan B. Goldman and Ricardo Martin-Brualla and Steven M. Seitz},
  title     = {HyperNeRF: A Higher-Dimensional Representation for Topologically Varying Neural Radiance Fields},
  booktitle = CVPR,
  pages     = {8151--8161},
  year      = 2021
}

@inproceedings{Pumarola21_DNeRF,
  author    = {Adri{\`a} Pumarola and Enric Corona and Gerard Pons-Moll and Javier Romero and Francesc Moreno-Noguer},
  title     = {{D-NeRF}: Neural Radiance Fields for Dynamic Scenes},
  booktitle = CVPR,
  pages     = {10318--10327},
  year      = 2021
}

@inproceedings{Chen22_Tensorf,
  author    = {Anpei Chen and Zhang, Zexiang and Wang, G. and Ding, R. and Liu, X. and Zhang, J. and Yu, J.},
  title     = {{TensoRF}: Tensorial Radiance Fields},
  booktitle = ECCV,
  pages     = {272--289},
  year      = 2022
}

@article{Cao2023HEXPLANE,
    author    = {Cao, Ang and Johnson, Justin},
    title     = {HexPlane: A Fast Representation for Dynamic Scenes},
    journal   = {CVPR},
    year      = {2023},
    }

@article{feng2024gaussian,
  title   = {Gaussian splashing: Dynamic fluid synthesis with gaussian splatting},
  author  = {Feng, Yutao and Feng, Xiang and Shang, Yintong and Jiang, Ying and Yu, Chang and Zong, Zeshun and Shao, Tianjia and Wu, Hongzhi and Zhou, Kun and Jiang, Chenfanfu and others},
  journal = {arXiv preprint arXiv:2401.15318},
  year    = {2024}
}

@article{kerbl20233d,
  title     = {3d gaussian splatting for real-time radiance field rendering},
  author    = {Kerbl, Bernhard and Kopanas, Georgios and Leimk{\"u}hler, Thomas and Drettakis, George},
  journal   = {ACM Transactions on Graphics},
  volume    = {42},
  number    = {4},
  pages     = {1--14},
  year      = {2023},
  publisher = {ACM}
}

@inproceedings{kplanes_2023,
  title     = {K-Planes: Explicit Radiance Fields in Space, Time, and Appearance},
  author    = {{Sara Fridovich-Keil and Giacomo Meanti} and Frederik Rahbæk Warburg and Benjamin Recht and Angjoo Kanazawa},
  year      = {2023},
  booktitle = {CVPR}
}

@inproceedings{li2022neural,
  title     = {Neural 3d video synthesis from multi-view video},
  author    = {Li, Tianye and Slavcheva, Mira and Zollhoefer, Michael and Green, Simon and Lassner, Christoph and Kim, Changil and Schmidt, Tanner and Lovegrove, Steven and Goesele, Michael and Newcombe, Richard and others},
  booktitle = cvpr,
  year      = {2022}
}

@inproceedings{lin2024gaussian,
  title     = {Gaussian-flow: 4d reconstruction with dynamic 3d gaussian particle},
  author    = {Lin, Youtian and Dai, Zuozhuo and Zhu, Siyu and Yao, Yao},
  booktitle = {Proceedings of the IEEE/CVF Conference on Computer Vision and Pattern Recognition},
  pages     = {21136--21145},
  year      = {2024}
}

@inproceedings{luiten2024dynamic,
  title        = {Dynamic 3d gaussians: Tracking by persistent dynamic view synthesis},
  author       = {Luiten, Jonathon and Kopanas, Georgios and Leibe, Bastian and Ramanan, Deva},
  booktitle    = {2024 International Conference on 3D Vision (3DV)},
  pages        = {800--809},
  year         = {2024},
  organization = {IEEE}
}

@article{mildenhall2020nerf,
  title   = {NeRF: Representing Scenes as Neural Radiance Fields for View Synthesis},
  author  = {Mildenhall, Ben and Srinivasan, Pratul P and Tancik, Matthew and Barron, Jonathan T and Ramamoorthi, Ravi and Ng, Ren},
  journal = {arxiv}, 
  year    = {2020}
}

@inproceedings{nerf,
  title     = {{NeRF}: Representing Scenes as Neural Radiance Fields for View Synthesis},
  author    = {Ben Mildenhall and Pratul P. Srinivasan and Matthew Tancik and Jonathan T. Barron and Ravi Ramamoorthi and Ren Ng},
  year      = {2020},
  booktitle = {ECCV}
}

@article{park2021hypernerf,
  title   = {Hypernerf: A higher-dimensional representation for topologically varying neural radiance fields},
  author  = {Park, Keunhong and Sinha, Utkarsh and Hedman, Peter and Barron, Jonathan T and Bouaziz, Sofien and Goldman, Dan B and Martin-Brualla, Ricardo and Seitz, Steven M},
  journal = {arXiv},
  year    = {2021}
}

@inproceedings{park2021nerfies,
  title     = {Nerfies: Deformable neural radiance fields},
  author    = {Park, Keunhong and Sinha, Utkarsh and Barron, Jonathan T and Bouaziz, Sofien and Goldman, Dan B and Seitz, Steven M and Martin-Brualla, Ricardo},
  booktitle = cvpr,
  year      = {2021}
}

@inproceedings{pumarola2021d,
  title     = {D-nerf: Neural radiance fields for dynamic scenes},
  author    = {Pumarola, Albert and Corona, Enric and Pons-Moll, Gerard and Moreno-Noguer, Francesc},
  booktitle = cvpr,
  year      = {2021}
}

@inproceedings{zhang2018unreasonable,
  title     = {The Unreasonable Effectiveness of Deep Features as a Perceptual Metric},
  author    = {Zhang, Richard and Isola, Phillip and Efros, Alexei A and Shechtman, Eli and Wang, Oliver},
  booktitle = cvpr,
  year      = {2018}
}

@article{Kerbl2023_3DGaussianSplatting,
  author    = {Bernhard Kerbl and Georgios Kopanas and Thomas Leimk{\"u}hler and George Drettakis},
  title     = {3D Gaussian Splatting for Real-Time Radiance Field Rendering},
  journal   = {ACM Transactions on Graphics},
  volume    = {42},
  number    = {4},
  year      = {2023},
  url       = {https://repo-sam.inria.fr/fungraph/3d-gaussian-splatting/}
}

@inproceedings{Wu2024_4DGaussianSplatting,
  author    = {Guanjun Wu and Taoran Yi and Jiemin Fang and Lingxi Xie and Xiaopeng Zhang and Wei Wei and Wenyu Liu and Qi Tian and Xinggang Wang},
  title     = {4D Gaussian Splatting for Real-Time Dynamic Scene Rendering},
  booktitle = {Proceedings of the IEEE/CVF Conference on Computer Vision and Pattern Recognition (CVPR)},
  year      = {2024},
  note      = {arXiv preprint arXiv:2310.08528},
  url       = {https://openaccess.thecvf.com/content/CVPR2024/html/Wu_4D_Gaussian_Splatting_for_Real-Time_Dynamic_Scene_Rendering_CVPR_2024_paper.html}
}

@misc{Zhang2024_MEGA,
  author    = {Xinjie Zhang and Zhening Liu and Yifan Zhang and Xingtong Ge and Dailan He and Tongda Xu and Yan Wang and Zehong Lin and Shuicheng Yan and Jun Zhang},
  title     = {MEGA: Memory-Efficient 4D Gaussian Splatting for Dynamic Scenes},
  howpublished = {arXiv preprint arXiv:2410.13613},
  year      = {2024},
  note      = {10.48550/arXiv.2410.13613},
  url       = {https://arxiv.org/abs/2410.13613}
}

@article{yang2023deformable3dgs,
    title={Deformable 3D Gaussians for High-Fidelity Monocular Dynamic Scene Reconstruction},
    author={Yang, Ziyi and Gao, Xinyu and Zhou, Wen and Jiao, Shaohui and Zhang, Yuqing and Jin, Xiaogang},
    journal={arXiv preprint arXiv:2309.13101},
    year={2023}
}

@inproceedings{Terzopoulos1987_ElasticallyDeformableModels,
  author    = {Demetri Terzopoulos and John Platt and Alan Barr and Kurt Fleischer},
  title     = {Elastically Deformable Models},
  booktitle = {Computer Graphics (Proceedings of SIGGRAPH ’87)},
  volume    = {21},
  number    = {4},
  pages     = {205--214},
  year      = {1987}
}

@inproceedings{Stam1999StableFluids,
  author    = {Jos Stam},
  title     = {Stable Fluids},
  booktitle = {Proceedings of SIGGRAPH ’99},
  pages     = {121--128},
  year      = {1999}
}

@book{Bridson2008_FluidSimulation,
  author    = {Robert Bridson},
  title     = {Fluid Simulation for Computer Graphics},
  publisher = {A K Peters},
  year      = {2008}
}

@article{Gregson2014_CaptureFluid,
  author    = {James Gregson and Ivo Ihrke and Wolfgang Heidrich},
  title     = {From Capture to Simulation: Connecting Fluid Reconstruction and Simulation},
  journal   = {ACM Transactions on Graphics (TOG)},
  volume    = {33},
  number    = {4},
  pages     = {1--11},
  year      = {2014}
}

@inproceedings{Okabe2015_Fluid,
  author    = {Makoto Okabe and Yasuyuki Matsushita and Takeo Igarashi},
  title     = {Fluid Volume Reconstruction from Multi-view Video},
  booktitle = {IEEE International Conference on Computer Vision (ICCV)},
  year      = {2015}
}

@article{Eckert2019_ScalarFlow,
  author    = {Michael Eckert and Nils Thuerey},
  title     = {ScalarFlow: A Large-Scale Volumetric Data Set of Real-World Scalar Transport Flows for Computer Animation and Machine Learning},
  journal   = {ACM Transactions on Graphics (TOG)},
  volume    = {38},
  number    = {4},
  pages     = {1--15},
  year      = {2019}
}

@article{Kim2019_DeepFluids,
  author    = {Byungsoo Kim and Vinicius C. Azevedo and Markus Gross and Nils Thuerey},
  title     = {Deep Fluids: A Generative Network for Parameterized Fluid Simulations},
  journal   = {Computer Graphics Forum (Eurographics)},
  volume    = {38},
  number    = {2},
  pages     = {59--70},
  year      = {2019}
}

@inproceedings{Wiewel2019_LatentSpaceFluids,
  author    = {Stefan Wiewel and Byungsoo Kim and Nils Thuerey},
  title     = {Latent Space Physics: Towards Learning the Temporal Evolution of Fluid Simulations},
  booktitle = {Computer Graphics Forum (Eurographics)},
  volume    = {38},
  number    = {2},
  pages     = {71--82},
  year      = {2019}
}

@article{Franz2021_GlobalNeural,
  author    = {Ernst Franz and Nils Thuerey},
  title     = {Global Neural Flow: Learning Generalizable Fluid Dynamics from Visual Data},
  journal   = {ACM Transactions on Graphics (TOG)},
  volume    = {40},
  number    = {6},
  pages     = {1--14},
  year      = {2021}
}

@article{Deng2023_NeuralFlowMaps,
  author    = {Yitong Deng and Hong-Xing Yu and Diyang Zhang and Jiajun Wu and Bo Zhu},
  title     = {Fluid Simulation on Neural Flow Maps},
  journal   = {ACM Transactions on Graphics (TOG)},
  volume    = {42},
  number    = {6},
  pages     = {244:1--244:15},
  year      = {2023}
}

@article{FluidNexus2024,
  author    = {H. Zhang and X. Liu and Y. Gao and Y. Wang and B. Zhu},
  title     = {FluidNexus: Neural Video-based Fluid Reconstruction and Prediction},
  journal   = {arXiv preprint arXiv:2404.01563},
  year      = {2024}
}

@inproceedings{Chen2018NeuralODE,
  author    = {Ricky T. Q. Chen and Yulia Rubanova and Jesse Bettencourt and David Duvenaud},
  title     = {Neural Ordinary Differential Equations},
  booktitle = {Advances in Neural Information Processing Systems (NeurIPS)},
  year      = {2018}
}

@article{Kochkov2021_MLAcceleratedCFD,
  author    = {Dmitry Kochkov and Amit Maity and Max Zwicker and Nils Thuerey and Justin Knoll},
  title     = {Machine Learning--Accelerated Computational Fluid Dynamics},
  journal   = {Proceedings of the National Academy of Sciences},
  volume    = {118},
  number    = {20},
  pages     = {e2101784118},
  year      = {2021}
}

@inproceedings{Schenck2018_SPNets_DifferentiableFluidDynamics,
  author    = {Connor Schenck and Dieter Fox},
  title     = {SPNets: Differentiable Fluid Dynamics for Deep Neural Networks},
  booktitle = {arXiv preprint arXiv:1806.06094},
  year      = {2018}
}

@article{ElHassan2025_PINNsFluidDynamics,
  author    = {Mouhammad El Hassan and Ali Mjalled and Philippe Miron and Martin M{\"o}nnigmann and Nikolay Bukharin},
  title     = {Machine Learning in Fluid Dynamics---Physics-Informed Neural Networks (PINNs) Using Sparse Data},
  journal   = {Fluids},
  volume    = {10},
  number    = {9},
  pages     = {226},
  year      = {2025}
}

@inproceedings{kwon2025efficient,
  title     = {Efficient Editable 4D Gaussian Fields for Dynamic Scene Rendering},
  author    = {Kwon, Youngjoong and Kim, Minhyuk and Kim, Seungyong and Park, Jeong Joon and Choi, Jonghyun and Kim, Jaesik},
  booktitle = {Proceedings of the IEEE/CVF Conference on Computer Vision and Pattern Recognition (CVPR)},
  year      = {2025}
}

@inproceedings{lee2024fully,
  title     = {Fully Explicit Dynamic Gaussian Splatting for Real-Time Dynamic View Synthesis},
  author    = {Lee, Jung-Woo and Kim, Jae-Han and Kim, Jaesik},
  booktitle = {European Conference on Computer Vision (ECCV)},
  year      = {2024}
}

@inproceedings{Zhu2024_MotionGS,
  author    = {Ruijie Zhu and Yanzhe Liang and Hanzhi Chang and Jiacheng Deng and Jiahao Lu and Wenfei Yang and Tianzhu Zhang and Yongdong Zhang},
  title     = {MotionGS: Exploring Explicit Motion Guidance for Deformable 3D Gaussian Splatting},
  booktitle = {Advances in Neural Information Processing Systems (NeurIPS) 2024},
  year      = {2024}
}

@inproceedings{Kratimenos2024_DynMF,
  author    = {Agelos Kratimenos and Jiahui Lei and Kostas Daniilidis},
  title     = {DynMF: Neural Motion Factorization for Real-time Dynamic View Synthesis with 3D Gaussian Splatting},
  booktitle = {European Conference on Computer Vision (ECCV) 2024},
  year      = {2024}
}

@article{Hu2025_MotionDecoupled3DGS,
  author    = {X Hu and others},
  title     = {Motion Decoupled 3D Gaussian Splatting for Dynamic Object Representation with Large Motion from a Monocular Camera},
  journal   = {AAAI Conference on Artificial Intelligence (AAAI) 2025},
  year      = {2025}
}

@article{Guo2024_MotionAware3DGS,
  author    = {Zhiyang Guo and Wengang Zhou and Li Li and Min Wang and Houqiang Li},
  title     = {Motion‐aware 3D Gaussian Splatting for Efficient Dynamic Scene Reconstruction},
  journal   = {arXiv preprint arXiv:2403.11447},
  year      = {2024}
}

@article{Lee2025_CoMoGaussian,
  author    = {Jungho Lee and Donghyeong Kim and Dogyoon Lee and Suhwan Cho and Minhyeok Lee and Wonjoon Lee and Taeoh Kim and Dongyoon Wee and Sangyoun Lee},
  title     = {CoMoGaussian: Continuous Motion‐Aware Gaussian Splatting from Motion-Blurred Images},
  journal   = {arXiv preprint arXiv:2503.05332},
  year      = {2025}
}

@inproceedings{Xie_2024_CVPR,
  author    = {Tianyi Xie and Zeshun Zong and Yuxing Qiu and Xuan Li and Yutao Feng and Yin Yang and Chenfanfu Jiang},
  title     = {PhysGaussian: Physics-Integrated 3D Gaussians for Generative Dynamics},
  booktitle = {Proceedings of the IEEE/CVF Conference on Computer Vision and Pattern Recognition (CVPR)},
  year      = {2024},
  pages     = {4389--4398}
}

@inproceedings{Lin2024GaussianFlow,
  author    = {Youtian Lin and Zuozhuo Dai and Siyu Zhu and Yao Yao},
  title     = {Gaussian-Flow: 4D Reconstruction with Dynamic 3D Gaussian Particle},
  booktitle = {Proceedings of the IEEE/CVF Conference on Computer Vision and Pattern Recognition (CVPR)},
  year      = {2024},
  pages     = {21136--21145}
}

@article{SplatFlow2024,
  author    = {Chengyang Xie and Qiang Liu and Xinyue Zhang and Wenhao Xu and Jianfeng He and Yifan Liu},
  title     = {SplatFlow: Self-Supervised Scene Flow Estimation with 3D Gaussian Splatting},
  journal   = {arXiv preprint arXiv:2410.12345},
  year      = {2024}
}

@InProceedings{Li_2022_CVPR,
  author    = {Li, Tianye and Slavcheva, Mira and Zollhöfer, Michael and Green, Simon and Lassner, Christoph and Kim, Changil and Schmidt, Tanner and Lovegrove, Steven and Goesele, Michael and Newcombe, Richard and Lv, Zhaoyang},
  title     = {Neural 3D Video Synthesis From Multi-View Video},
  booktitle = {Proceedings of the IEEE/CVF Conference on Computer Vision and Pattern Recognition (CVPR)},
  month     = {June},
  year      = {2022},
  pages     = {5521-5531}
}

@article{cen2023saga,
      title={Segment Any 3D Gaussians}, 
      author={Jiazhong Cen and Jiemin Fang and Chen Yang and Lingxi Xie and Xiaopeng Zhang and Wei Shen and Qi Tian},
      year={2023},
      journal={arXiv preprint arXiv:2312.00860},
}

@misc{liu2023zero1to3,
      title={Zero-1-to-3: Zero-shot One Image to 3D Object}, 
      author={Ruoshi Liu and Rundi Wu and Basile Van Hoorick and Pavel Tokmakov and Sergey Zakharov and Carl Vondrick},
      year={2023},
      eprint={2303.11328},
      archivePrefix={arXiv},
      primaryClass={cs.CV}
}

@article{bruce2024genie,
  title={GENIE: Generative Interactive Environments},
  author={Bruce, J. and Schrittwieser, J. and Mirza, M. and others},
  journal={arXiv preprint arXiv:2402.15329},
  year={2024}
}

@misc{zhang2024recapturegenerativevideocamera,
      title={ReCapture: Generative Video Camera Controls for User-Provided Videos using Masked Video Fine-Tuning}, 
      author={David Junhao Zhang and Roni Paiss and Shiran Zada and Nikhil Karnad and David E. Jacobs and Yael Pritch and Inbar Mosseri and Mike Zheng Shou and Neal Wadhwa and Nataniel Ruiz},
      year={2024},
      eprint={2411.05003},
      archivePrefix={arXiv},
      primaryClass={cs.CV},
      url={https://arxiv.org/abs/2411.05003}, 
}

@article{ha2018worldmodels,
  title   = {World Models},
  author  = {Ha, David and Schmidhuber, J{\"u}rgen},
  journal = {arXiv preprint arXiv:1803.10122},
  year    = {2018}
}

@inproceedings{duisterhof2024deformgs,
  title     = {DeformGS: Scene Flow in Highly Deformable Scenes for Deformable Object Manipulation},
  author    = {Bardienus P. Duisterhof and Zhao Mandi and Yunchao Yao and Jia-Wei Liu and Jenny Seidenschwarz and Mike Zheng Shou and Deva Ramanan and Shuran Song and Stan Birchfield and Bowen Wen and Jeffrey Ichnowski},
  booktitle = {Proceedings of the 16th International Workshop on the Algorithmic Foundations of Robotics (WAFR)},
  year      = {2024}
}

@article{huang2023sc,
        title={SC-GS: Sparse-Controlled Gaussian Splatting for Editable Dynamic Scenes},
        author={Huang, Yi-Hua and Sun, Yang-Tian and Yang, Ziyi and Lyu, Xiaoyang and Cao, Yan-Pei and Qi, Xiaojuan},
        journal={arXiv preprint arXiv:2312.14937},
        year={2023}
      }

@inproceedings{luiten2023dynamic,
  title={Dynamic 3D Gaussians: Tracking by Persistent Dynamic View Synthesis},
  author={Luiten, Jonathon and Kopanas, Georgios and Leibe, Bastian and Ramanan, Deva},
  booktitle={3DV},
  year={2024}
}

@article{kalman1960new,
  title={A New Approach to Linear Filtering and Prediction Problems},
  author={Kalman, Rudolph Emil},
  journal={Journal of Basic Engineering},
  volume={82},
  number={1},
  pages={35--45},
  year={1960},
  publisher={ASME}
}

@article{kalman1961new,
  title={New Results in Linear Filtering and Prediction Theory},
  author={Kalman, Rudolph E. and Bucy, Richard S.},
  journal={Journal of Basic Engineering},
  volume={83},
  number={1},
  pages={95--108},
  year={1961},
  publisher={ASME}
}
